# Hybrid Q-Learning Applied to Ubiquitous recommender system

D. Bouneffouf
Télécom SudParis
9, rue Charles Fourier
91011 Evry, France
Djallel.Bouneffouf@it-sudparis.eu

## ABSTRACT

Ubiquitous information access becomes more and more important nowadays and research is aimed at making it adapted to users. Our work consists in applying machine learning techniques in order to bring a solution to some of the problems concerning the acceptance of the system by users.

To achieve this, we propose a fundamental shift in terms of how we model the learning of recommender system: inspired by models of human reasoning developed in robotic, we combine reinforcement learning and case-base reasoning to define a recommendation process that uses these two approaches for generating recommendations on different context dimensions (social, temporal, geographic). We describe an implementation of the recommender system based on this framework. We also present preliminary results from experiments with the system and show how our approach increases the recommendation quality.

## Categories and Subject Descriptors

H.3.3 [**Information Search and Retrieval**]: *information filtering, Selection process, Relevance feedback* .

## General Terms

Algorithms

## 1. INTRODUCTION

The need for adapting information systems to the user context has been accentuated by the extensive development of mobile applications that provide a considerable amount of data of all types (images, texts, sounds, videos, etc.). It becomes thus crucial to help users by guiding them in their access to information.

Systems should be able to recommend information helping the user to fulfill his/her goal. The information given by the system depends on the user's situation, i. e. an instance of the context. Possible situations and the associated actions reflect the user's work habits.

Major difficulties when applying techniques to adapt a system to the user follow:

- **Avoiding the intervention of experts**: on one hand, experts are not sure of the interest of the user, may define wrong ideas about him; on the other hand, an expert is not always available.

- **Starting from scratch:** in the initial state, the system's behavior should not be incoherent for the user to not refuse it quickly.

- **A slow learning process:** the learning process has to be quick to avoid bothering the user with incorrect recommendation.

-**The evolution of the user's interest**: the interest of the user may change with the time. The system has to be continuously adapted to this dynamic change using the user's context information to provide the relevant recommendations because, if the system behavior is incoherent, the user refuses it quickly.

We sum up all of these problems in the following scenario.
Senario. Given the company Nomalys, the set of marketing staff people can access to the most relevant data of their company via their mobile phone. Paul is a new sales representative of the company; he is integrating a team of ten marketing staff members. Our recommender system has to retrieve the relevant information to this user to help him for doing his job.

To solve the problem of the scenario, our recommender system has to retrieve information about the user and his context from his mobile device that user brings into the environment. The system uses the context knowledge to propose relevant information to the user. For instance, regarding Paul's agenda, Paul has a meeting with a client in Paris at midday. When he arrives at his meeting, the system should recommend him the client's register of complaints, which would help Paul to better manage his meeting.

Our system starts with a predefined set of actions defined by the user's social group and adapts it progressively to a particular user. This default behavior allows the system to be ready-to-use and the learning is a lifelong process. Thus, the system will, at first, be only acceptable to the user, and will, as time passes, give more and more satisfying results.

In summary, the recommender system observes the user and gets information from his context and his activity. For that it needs perceptive sensor modules capable of providing this kind of information. Our ubiquitous system is composed of sensor modules which can fire events, received by the recommender system. This input allows the recommender system to estimate the user's situation. The default behavior, possibly modified by acquired experience, indicates to the recommender system how to act in a certain situation. When the appropriate action is chosen, the recommender system executes it.

In the remaining of this paper, Section 2 is dedicated to the state of the art. Then, in section 3, we describe the current ideas of our ongoing work, followed by results in Section 4. Finally, we conclude, giving directions for future work.

## 2. State of the art

The trend today on recommendation systems is to recommend relevant information to users, using supervised machine learning techniques. In that type of techniques, the recommender system has to pass by two steps: (1) The learning step, where examples are presented to the system; which "learns" from examples and gradually adjusts its parameters to the desired output. (2)Exploitation step: new examples never seen before are presented to the system and ask it for generalizing [10].

These approaches have good results. However, they need an amount of experience provided by an expert. They cannot start from scratch and they are slow. Moreover, the user's interest can change with the time, and the techniques cannot really follow this. Some works found in literature try to solve those problems, as explained in what follows.

-**Starting from scratch:** to avoid this problem, which is commune to machine learning algorithms, in [7] authors use collaborative filtering to consider demographic information about users for providing them more accurate prediction, but their system does not follow the user's interest evolution.

- **Avoiding the intervention of experts**: To avoid the intervention of an expert, in [9] the authors use Reinforcement Learning (RL), which is a good alternative because it does not need a previous experience to start work. However, a major difficulty when applying RL techniques to real world problems is their slow convergence.

-**Accelerate the learning process:** In [9], the author proposes to accelerate RL by using indirect Q-learning. However, their recommendation system starts with a set of actions which are predefined by them.

-**The evolution of the user's interest**: The authors on [19] propose to follow the interest of the user by using an exploration strategy on the q-learning algorithm. But they don't care about the others problems cited above.

We can observe that each work cited above tries to solve only one of those problems and none of them proposes to solve all of them at the same time.

To create a system avoiding all the problems, we propose to use the Q-learning algorithm with an exploration strategy to solve the problem of intervention of an expert and follow the user's interest evolution. For the starting from scratch problem, we give Q-learning algorithm the ability to explore the knowledge of other users by using collaborative filtering. To accelerate the Q-learning process, we mix it with case base reasoning techniques to allow the reuse of the case-base and satisfy the user more quickly. We were inspired by case base reasoning to accelerate reinforcement learning techniques introduced and implemented by [11] in robotic.

## 3. Proposition

### 3.1 Reinforcement learning and the Q-learning algorithm

The goal of the agent in a RL problem is to learn an optimal policy $\pi^*: S \rightarrow A$ that maps the current state s into the most desirable action *a* to be performed in *s*.

One strategy to learn the optimal policy $\pi^*$ is to allow the agent to learn the evaluation function $Q: S \times A \rightarrow R$. Each action value $Q(s, a)$ represents the expected cost incurred by the agent when taking action *a* at state *s* and following an optimal policy thereafter.

The Q–learning algorithm [14] is a well-know RL technique that uses a strategy to learn an optimal policy $\pi^*$ via learning of the action values. It iteratively approximates Q, provided the system can be modeled as a Markov decision process (MDP), the reinforcement function is bounded, and actions are chosen so that every state-action pair can visit an infinite number of times. The Q-learning update rule is:

$$Q(s, a) \leftarrow Q(s, a) + \alpha[r + \gamma \max_a Q(s', a') - Q(s, a)] , \quad (1)$$

where s is the current state; **a** is the action performed in **s**; **r** is the reward received; **s′** is the new state; $\gamma$ is the discount factor ($0 \leq \gamma < 1$); and **α** is the learning rate.

### 3.2 Collaborative filtering

A Collaborative Filtering (CF) recommender system works as follows. Given a set of transactions *D*, where each transaction *T* is of the form <*id*, *item*, *rating*>, a recommender model *M* is produced. Each item is represented by a categorical value, while the rating is a numerical value in a given scale (e.g. each item is a movie rated with 1 to 5 stars). Such a model *M* can produce a list of *top-N* recommended items, and corresponding predicted ratings, from a given set of known ratings [4]. In many situations, ratings are not explicit. For example, if we want to recommend Web pages to a Web site visitor, we can use the set of pages she or he has visited, assigning those pages an implicit rate of one, and zero to all the other pages.

In terms of CF, three major classes of algorithms exist: *Memory-based, Model-based* and *Hybrid-based* [1, 4]. At the moment, in our work, we use the simplest of them which is the memory-based CF. In memory-based CF, the whole set of transactions is stored and is used by the recommender model. These algorithms employ a notion of distance to find a set of users, known as neighbors, who tend to agree with the target user. The preferences of neighbors are then combined to produce a prediction or *top-N* recommendation for the active user.

### 3.3 Case Based Reasoning

Case based reasoning (CBR) [12, 13] uses knowledge of previous situations (cases) to solve new problems, by finding a similar past case and reusing it in the new problem situation.

According to Lopez de Mantaras *et al* [12], solving a problem by CBR involves "obtaining a problem description, measuring the similarity of the current problem to previous problems stored in a case base with their known solutions, retrieving one or more similar cases, and attempting to reuse the solution of the retrieved case(s), possibly after adapting it to account for differences in problem descriptions". This is some works found in the literature which use this technique [17, 18].

### 3.4 The hybrid Q-learning (HyQL)

We improve the performance of the Q-learning, in the following point:

-**Reuse case:** To accelerate the Q-learning algorithm, we propose to integrate CBR in the loop of the Q-learning algorithm. For each step of Q-learning, before choosing the best action, the algorithm computes the similarity and, if there is a case that can be reused, the algorithm retrieves and adapts it.

-**Using social group:** In the Q-Learning algorithm, it is said that, for every state *s*, action $a = Q(s)$ is chosen according to the current policy. The choice of the action by the policy must ensure a balance between exploration and exploitation.

The exploitation is to choose the best action for the current state, thus exploiting the system's knowledge. The exploration is to choose an action other than the best one in order to test it, observe its consequences, and increase the knowledge of the system.

There are several strategies to make the balance between exploration and exploitation. Here, we focus on two of them: the **greedy** strategy chooses always the best action from the Q-table; the **ε-greedy** strategy adds some greedy exploration policy, choosing a random action at each step if the policy returns the greedy action (probability ε) or a random action (probability *1 - ε)*.

To give the Q-Learning the ability to use advices from other users sharing the same ideas, we propose to extend the -ε-greedy strategy of the Q-Learning algorithm with the ability to explore the knowledge of other users. In the -ε-greedy strategy of the exploration/exploitation functions, we replace the random action by an action that is selected by calculating the similarity of user profiles applying the CF algorithm. The equation 2 shows how it is done.

$$\pi(s) = \begin{cases} \text{argmax}_a\ Q(s, a) & \text{if } q \leq p, \\ a_{\text{users advises}} & \text{otherwise} \end{cases} \quad (2)$$

In equation 2:
– **q** is a random value uniformly distributed over [0, 1] and **p** ($0 \leq$ **p** $\leq 1$) is a parameter that defines the exploration/exploitation tradeoff: the larger is **p**, the smaller is the probability of executing a random exploratory action.
– **a** $_{\text{users advises}}$ is an action chosen among those available in state **s** by applying the CF algorithm.

The complete proposed hybrid Q-learning algorithm, called HyQL algorithm follows.

**The HyQL algorithm:**
   Initialize Qt(**s**, **a**) arbitrarily.
   Repeat (for each episode):
     Initialize s.
     Repeat (for each step):
       Compute similarity and cost.
       If there is a case that can be reused:
         Retrieve and adapt if necessary.
       Select an action **a** using equation 2.
       Execute the action **a**, observe r(**s**, **a**), **s′**.
       Update the values of Q(**s**, **a**) according to equation 1.
       **s ← s′**.
     Until **s** is terminal.
   Until some stopping criterion is reached.

## 4. Global mechanism

To evaluate our algorithm we implement it in a ubiquitous recommender system. Figure 1 summarizes the global mechanism of the recommender system. To detect the user's context, the recommender system receives events from the sensor module. These events constitute the input of the recommendation system and launch the reasoning module. Based on this input, the reasoning module allows choosing an action to be executed in the environment.

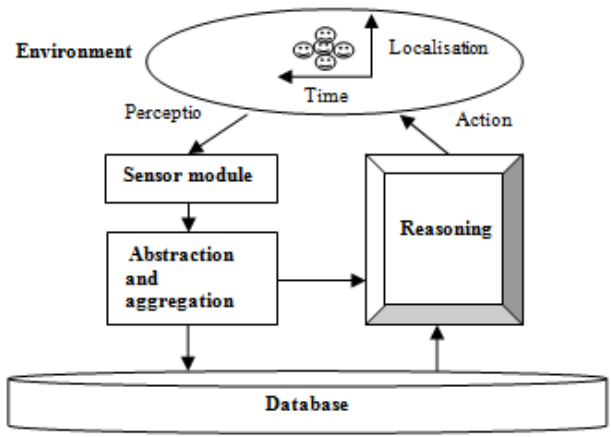

**Figure 1:** Global mechanism of the interaction between system and environment

### 4.1 Environment
We consider the environment being composed of all context dimensions described in [5], namely cognitive, time, geographic, material, social, source document, qualitative, psychological, and demographic.

### 4.2 Sensor module
In our work, the sensor module detects time, geographic, social, and cognitive dimensions of the context in the following way:

(1) The cognitive dimension is given by all the actions of the user, like for example: navigation (reads a document, opens a folder, etc…), sending an email and calling.

(2) The social group is predefined by the user. We suppose for example that all the marketing users of the company have probably the same need in general, thus being part of the same group.

(3) The time is detected by the user mobile phone and the calendar of his/her company.

(4) The geographic dimension is detected by the GPS of the user.

### 4.3 Abstraction and aggregation
The abstraction is based on inference rules (e.g. specification / generalization) defined on the temporal or space ontology. For instance, if we consider the outputs of GPS, we use an operation of "reverse geocoding" to get the name and the type of the place.

The aggregation is the combination of the two dimensions of time and location, e.g. "morning at home." It allows more description of situations in various levels granularity.

To represent and characterize the location of the user, a model for the representation of geographical locations is required. To allow adequate representation of geographic information, the trend is towards semantic approaches with spatial ontology. As in [16], we propose to use ontology to represent and reason about geographic data.

To define the temporal aspects characterizing the situation of the user (morning, evening, weekend...), a clear model for representing and reasoning about time and time intervals is necessary.

To allow for an adequate representation of temporal information and its manipulation, the trend is towards semantic approaches with temporal ontology.

The OWL-Time ontology [15] is today a reference for representing and reasoning about time. We propose to base our work on this ontology and extend it if necessary.

### 4.4 Database
All modules of the system share a database divided into four parts: *user*, *Preferences*, *history* and *devices*.

The *history* part stores all occurred events and all actions taken by the system. This part is useful for inferring the good recommendation to the user and it is divided into: **Action_history**, which contains all the interaction of the system with the environment; **Event_history**, which contains all the events registered by the user on his calendar.

The *devices* part contains information about the devices. This knowledge can be used to determine what the device can do.

The *Preferences* is a part with contains the aggregation of actions of the recommender system with the users' rewards.

The *user* part describes registered users (it stores user logins allowing identifying them).

### 4.5 Reasoning system
The reasoning system allows choosing an action to deliver on each situation. In our experiments, the reasoning module is controlled by each of the previously presented algorithms: CF, Q-Learning, CBR and HyQL.

## 5. CONCLUSION

The aim of this work is to investigate the problems that we find when we try to adapt a recommender system to the user in a ubiquitous environment. The recommender system defines the observable situations and what actions should be executed in each situation in order to provide useful information to the user.

To achieve this goal, we propose to mix the RL algorithm with CBR and CF algorithms. As future work, we intend to carry out tests with more users and case-base from Nomalys company.

## 6. ACKNOWLEDGMENTS

This work is partially funded by Nomalys French Company (www.nomalys.com).